\def\year{2021}\relax

\documentclass[letterpaper]{article} 

\usepackage{aaai21}  
\usepackage{times}  
\usepackage{helvet} 
\usepackage{courier}  
\usepackage[hyphens]{url}  
\usepackage{graphicx} 
\usepackage{natbib}  
\usepackage{caption} 
\usepackage[switch]{lineno} 
\usepackage{amsmath}%

\title{Cross-Modal Alignment with Mixture Experts Neural Network \\for Intral-City Retail Recommendation}
\author{
   Po Li,
   Lei Li,
   Yan Fu,
  Jun Rong, 
  Yu Zhang
\\
}
\affiliations{

 Alibaba Group, Hang Zhou, China\\
   \{poli.lp,zy223687,leili.ll,yanfu.fy,junrong.rj\}@alibaba-inc.com
}

\begin{document}
\maketitle

\begin{abstract}
In this paper, we introduce Cross-modal Alignment with mixture experts Neural Network (CameNN) recommendation model for intral-city retail industry, which aims to provide fresh foods and groceries retailing within 5 hours delivery service arising for the outbreak of Coronavirus disease (COVID-19) pandemic around the world. As most foods and groceries stories are small business without extra labor to type or maintain good commodities' properties information carefully in merchant inventory system. Inaccurate text information and vague image data are common issues in this industry. Pioneer Conversion Rate (CVR) prediction models from FM series models, Deep \& Wide Models to the state-of-art Transformer models mainly focus on how to learn more effective interaction of features. Consuming those misleading image and text data, implementation of above-mentioned CVR models are not able to achieve satisfying performance as they do in other industries where data is inaccurate. Meanwhile, numerous frustrating bad cases happen in our practical recommendation.To this end, we propose CameNN, which is a multi-task model with three tasks including Image to Text Alignment (ITA) task, Text to Image Alignment (TIA) task and CVR prediction task. We use pre-trained BERT to generate the text embedding and pre-trained InceptionV4 to generate image patch embedding (each image is split into small patches with the same pixels and treat each patch as an image token). Softmax gating networks follow to learn the weight of each transformer expert output and choose only a subset of experts conditioned on the input. Then transformer encoder is applied as the share-bottom layer to learn all input features' shared interaction. Next, mixture of transformer experts (MoE) layer is implemented to model different aspects of tasks. At top of the MoE layer, we deploy a transformer layer for each task as task tower to learn task-specific information. On the real word intra-city dataset, experiments demonstrate CameNN outperform baselines and achieve significant improvements on the image and text representation. In practice, we applied CameNN on CVR prediction in our intra-city recommender system which is one of the leading intra-city platforms operated in China.
\end{abstract}

\section{1  Introduction}
With the outbreak of Coronavirus disease (COVID-19) pandemic around the world, this newly discovered coronavirus has caused more than half millions of death and tens millions of confirmed cases. According to WHO report \cite{who:2020}, keep 'physical distancing'  is one of the most effective measures to prevent the spread of COVID-19. The COVID-19 pandemic totally changed the life of most people in this planet, such as most companies require their staffs to work from home instead of office and people intend to buy daily fresh foods and groceries through online stores instead of physical stores. Intra-city online retailing service, which provides daily necessaries for people within 5 hours or even less than 30 minutes, help people to prepare fresh foods and buy daily necessaries conveniently and timely. However, most groceries or fresh foods stores are family-owned business with few internet retailing experience and knowledge. Intra-city online retailing platform that providing service such as online stores and intra-city delivery for above-mentioned businesses has arisen up. Personalized goods recommender system plays vital role in shopping guide for customers to the right commodities. But for most physical groceries being small stories without extra staffs to type commodities' proprieties into online inventory system carefully, lots of information post on the online platform are incorrect such as put item name  'Cauliflower'  into category 'Fruit' , upload vague or wrong images for the item and even describe 'Strawberries' with item name 'Berries'. Those misleading item data consumed as inputs result in the poor performance of the item recommendation models. It's impossible to rectify millions of wrong data manually. One method to relieve is taking measures to reduce the wrong data generated from the origin source, which is time consuming. Another approach that designs a multi-task model to automatically rectify wrong data and achieve better commodities recommendation should be explored.

Recommender System mainly consists of two stage pipelines in our platform: matching and ranking. Stage one is matching, which is selecting a set of items according to users' profiles and behaviors. Then, use models to predict the Conversion Rate (CVR) for each item and rank them by certain rules. CVR prediction is a task to predict the probability of a user purchasing the recommended candidate items according to user's historical behaviors. In practical, models used in CVR prediction are similar with models of Click-through Rate (CTR) prediction that is predicting a user's click probability of candidate items based on user's historical behaviors.  The factorization machines (FMs) \cite{Steffen:2010:Factorization-Machines} is a typical CTR model designed to extend logistic regression (LR) with including second-order terms by allowing pairwise interactions between variables. Further extensions of FMs are also made to include field information (field-aware factorization machines (FFM); \cite{Yuchin:2016:Field-aware-Factorization-Machines}), attention mechanism (attention FM; \cite{Xiao:2017:Attentional-factorization-machines}), and even deep component (deep FM; \cite{Guo:2017:DeepFM}). However, other than deep FM, most FM-family models fail to include higher-order terms and the choice of second-order term requires domain expertise. Recently, deep learning models such as Wide \& Deep \cite{Cheng2016:Wide-deep},xDeepFM \cite{Lian:2018:xDeepFM}, deep interest network (DIN) \cite{guorui:2018:din} and Behavior Sequence Transformer (BST) \cite{qiwei:2019:bst} have been developed to learn the higher-order feature interactions. These deep models are bestowed with greater capacity of modeling user preference and capturing user behaviors. Whereas, for issues above-mentioned in intra-city retailing industry, those typical CTR models don't perform well as suffering from poor quality of text and image data. Numerous frustrating bad cases happen in our practical implementation. 

In this paper, we propose a multi-task recommendation model named Cross-modal Alignment with mixture experts Neural Network (CameNN) to solve the above problems. Three tasks including Text Alignment (ITA) task, Text to Image Alignment (TIA) task and CVR prediction task are introduced in CameNN. Inspired by FashionBERT \cite{dehong:2020:fashionbert}, we split each image into small patches with the same pixels and treat each patch as image token. Meanwhile, text tokens are tokenized according to \cite{yonghui:2016:wordpieces} into token sequences with adopting the standard BERT vocabulary. Pre-trained Chinese version BERT \cite{jacob:2018:bert} and pre-trained InceptionV4 \cite{christian:2016:inceptionv4} are implemented to generate text representation and image representation correspondingly. Then, we use mixture of transformer experts (MoE) layer to model different aspects of tasks with softmax gating networks which are designed to learn the weight of transformer experts and choose a subset of transformer experts conditioned on inputs. Next, shared-bottom layer is utilized to model input features' shared interaction for different tasks. At top of the MoE layer, we deploy a transformer layer for each task as task tower to learn task-specific information. Transformer encoder is the basic layer for share-bottom layers, expert layers and task tower layers. Experiments on dataset from a leading intra-city retailing platform operated in China demonstrate that CameNN outperform baselines on CVR prediction task and achieve significant improvements on rectifying image and text representation. In our real-world online application, CameNN do help to improve the CVR on items recommendation. \\

To summarize, our main contributions are as following:
\begin{itemize}
\item We describe the issues of intral-city retail recommendation facing and propose CameNN to address these difficulties. 

\item We present CameNN to conduct both cross-modal (images and text) alignment and customer Conversion Rate (CVR) prediction tasks on the real intral-city retail industry data to show that CameNN outperform baselines models on CRV task, TIA task and ITA task. 

\item We implement the ablation study to show the benefit of CameNN on text and image data alignment.

\item We show a successful and efficient large-scale online application of CameNN to improve CVR of items recommendation. 
\end{itemize}

\section{2  Related Work}

Traditional non-linear CTR models such as factorization machines (FMs) \cite{Steffen:2010:Factorization-Machines}  has been proven to be effective in recommendation systems. However, its modelling capacity is limited by its low complexity. To extend the ability of FM model, lots of efforts have been made, such as Field-aware FM (FFM) \cite{Yuchin:2016:Field-aware-Factorization-Machines} was proposed to learn different interaction with features from different fields and Attentional Factorization Machines (AFM)  \cite{Xiao:2017:Attentional-factorization-machines} was designed to use attention network  \cite{Bahdanau:2014:Attention} to learn the importance of each feature interaction. However, all these linear extensions of FM still focus on modeling the second-order feature interactions and impractical to deal with real-word non-linear structure data. 

In recent few years, benefiting from the booming of neural networks, models that are able to learn high-order features interaction has significantly improved the performance on CTR prediction. For example, Wide \& Deep \cite{Cheng2016:Wide-deep} model is designed to learn high-order feature interaction. Take advantage of higher-order feature learning of Wide \& Deep model and the second-order factorization power of FM \cite{Guo:2017:DeepFM}, an integration of Wide \& Deep model and FM model has been developed. Furthermore,extreme Deep factorization machine (xDeepFM) \cite{Lian:2018:xDeepFM} is established to exploit the modelling power of feed-forward neural networks. Ability to tackle sequential data is required to improve performance of CTR prediction, deep interest network (DIN) \cite{guorui:2018:din} and Behavior Sequence Transformer \cite{qiwei:2019:bst} are proposed to capture the dependencies between users' sequential behaviors that may reflect the interest behind historical behaviors. But above-mentioned models can't achieve good performance on CVR prediction with consuming misleading item data.  

For Multi-task learning models, a shared-bottom model structure is proposed by Caruana \cite{Caruana:1993:multitask} including bottom hidden layers shared across tasks. However, the task differences may cause optimization conflicts in this structure since all tasks share the same set of parameter of the shared-bottom layers. To avoid sharing same parameters, Duong et al. \cite{Duong:2015:lrdp} add L-2 constraints between two set of parameters for two tasks. Meanwhile, Misra et al. adapt the cross-stitch network  \cite{Misra:2016:csn} to learn a unique combination of task-specific embeddings and Yang et al. \cite{Yang:2016:dmrl} implement a tensor model to generate task-specific hidden parameters. All of these models require huge amounts of data to train and are inefficient for large-scale implementation. Ma et al. \cite{Ma:2018:mmoe} proposed Multi-gate Mixture-of-Experts (MMoE) model using softmax gating networks and mixture of identical multilayers perceptrons (MLP) with ReLU activations experts to accomplish multi-tasks more efficiently. To empower the model with ability to model user sequential behaviors, Zhen et al. \cite{zhen:2020:mose} developed a model named Multitask Mixture of Sequential Experts (MoSE) by replacing all the functional ReLu MLP layers in MMoE with LSTM layers \cite{Hochreiter:1997:LSTM} which can learn better sequential representations. However, MoSE did not explicitly handle cross-modality data such as images or text inputs. 

All aforementioned models need major modifications to tackle our issues in intra-city item recommendation. It inspires us to develop CameNN that can address this problem.

\section{3  Cross-modal Alignment with mixture experts Neural Network (CameNN)}

In this section, we give an overview of our proposed CameNN and describe the details how we use mixture experts to alignment text and image features and promote performance on the customer conversion rate(CVR).

\begin{figure*}[t]
\centering
\includegraphics[width=0.8\textwidth]{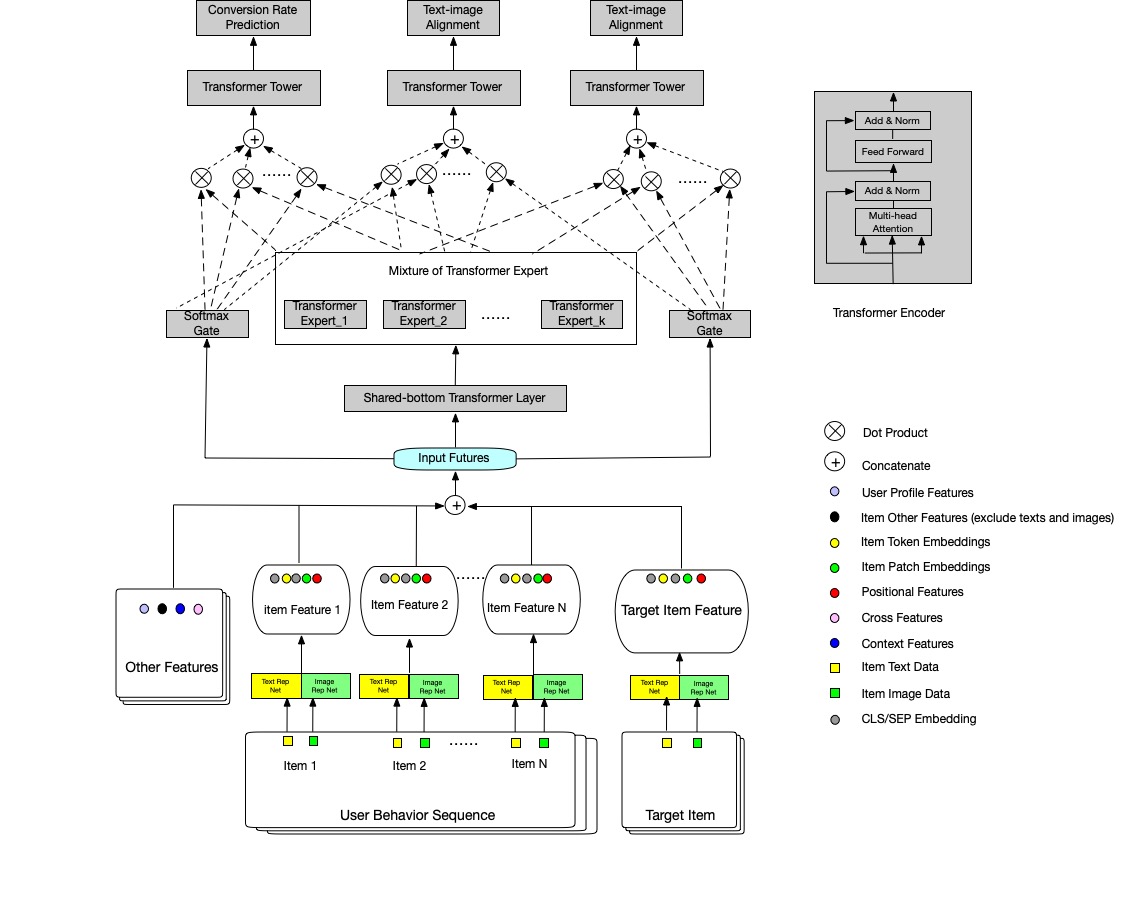} 
\caption{Our CameNN overview structure. The shared-bottom transformer layer, transformer experts and transformer towers are transformer encoder}.
\label{camenn-overview}
\end{figure*}

The overview structure of CameNN is depicted by Figure \ref{camenn-overview}. CameNN is composed of six parts: text representation net, image representation net, shared-bottom layer, softmax gating networks, mixture of experts and task tower net.

\begin{figure*}[t]
\centering
\includegraphics[width=0.8\textwidth]{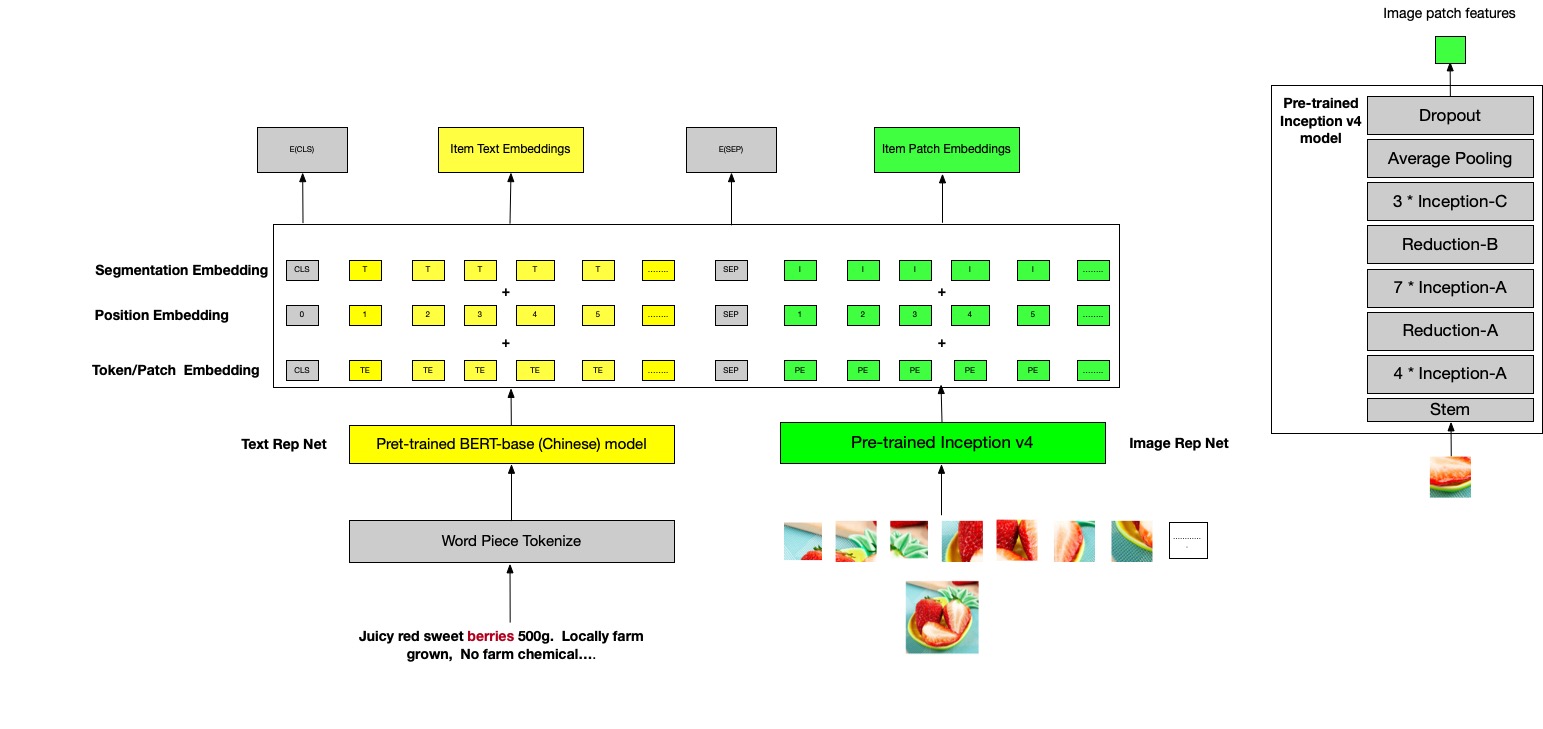} 
\caption{Use pre-trained Bert-base Chinese model to generate text embedding and pre-trained inception v4 to generate image patch embedding. The image is strawberry but the text information is vague}.
\label{tiembeding}
\end{figure*}

\subsection{3.1 Modeling Preliminary}
We use three categories of features: Other Features (including user profile features, item other features excluding texts and images, context Features, cross features), User Behavior Sequence (including list of items user bought with items' texts and images information) and Target item features(including items' texts and images information). For Other Features, they are encoded into one-hot vector as $x_o$. Then put $x_o$ through embedding layer to transform it into low-dimensional dense embedding as $E_{Other}$ with dimension size $n_E$.

\subsection{3.2 Text Representation Net}
BERT \cite{jacob:2018:bert} language model has been shown to be effective for various of natural language processing tasks. By using the standard prepossessing method of BERT, we tokenize the a item text $x_{txt}$ according to \cite{yonghui:2016:wordpieces} into token sequences with adopting the standard BERT vocabulary. Then we apply pre-trained BERT Chinese version with 12-layer ,768- hidden, 12-heads and 110M parameters to generate the token embedding with dimension size $n_E$. For segmentation, we use 'T' token for text token to differentiate text and image features as shown in Figure  \ref{tiembeding}. Meanwhile, special [CLS] tokens are placed in the first position of each item and [SEP] tokens between text tokens and image patch tokens. Next, similar to BERT, position embedding is to represent sequence position information. Then, sum the token embedding and position embedding as the final text input embedding representation. For the a item text embedding, we denote it as $e_{txt}$. The item text embedding generation would be:\\

 \begin{equation}
  e_{txt} = f_{BERT}(x_{txt}) + e_{poi\_t} + e_{seg\_t}\\
 \end{equation} 
where $e_{txt} \in R^{n_E \times n_T}$, $e_{poi\_t} \in R^{n_E \times n_T}$, $e_{seg\_t} \in R^{n_E \times n_T}$ , the $n_E$ is the embedding dimension size and $n_T$ is the numbers of split tokens in this item text. $e_{poi\_t}$ is the text tokens position embedding and $e_{seg\_t}$ is the text segmentation embedding.

\subsection{3.3 Image Representation Net}
The state-of-art method to extract semantic information from images is RoIs detection such as \cite{ross:2013:fastrnn} and use these RoIs as 'image tokens' which is shown in Figure  \ref{tiembeding}. In intra-city domain, item images are small and lack of detected RoIs. RoIs method is not effective in this industry. We apply another pioneer approach \cite{dehong:2020:fashionbert} to extract image patches as image tokens. To keep the rawer pixel feature than object-level RoIs of images, we split each image $x_{img}$ into small patches with the same pixels and treat each patch as an image token. Image patches are set in nature order patch sequences with non-repeated patches. 

Next, we implement a pre-trained image model InceptionV4 \cite{christian:2016:inceptionv4} as the main model to generate patch embedding with embedding dimension size of $n_E$. The pre-trained model could be any other pre-trained image model such as VGG-16 \cite{karen:2015:vgg16} or ResNeXt \cite{Saining:2017:resxnet}. Segmantion token 'I' is implemented for image patch tokens.  Place each patches in nature order, we represent the spatial information of patches by position embedding. Finally, image patches are represented as the sum of patch embedding, segmentation embedding and position embedding. The formulation is:

\begin{equation}
e_{img} = f_{InceptionV4}(x_{img}) + e_{poi\_p} + e_{seg\_p}
\end{equation}
where $e_{txt} \in R^{n_E \times n_P}$, $e_{poi\_p} \in R^{n_E \times n_P}$, $e_{seg\_p} \in R^{n_E \times n_P}$ the $n_E$ is the embedding dimension size and $n_P$ is the numbers of split patches in this item image.$e_{poi\_p}$ is the image patches position embedding and $e_{seg\_p}$ is the image segmentation embedding.

The input features $E_{Input}$ is generated as follows:

\begin{equation}
 \begin{split}
E_{Input} =& Concat(E_{Other},E_{User}, E_{Target}) \\
E_{User} =& [E_1, E_i,..E_N] \\
E_{Target} =& [e_{csl},e_{txt},e_{sep},e_{img}]\\
E_i =& [e_{csl,i},e_{txt,i},e_{sep,i},e_{img,i}]      
 \end{split}
\end{equation}
where  $e_{csl,i},e_{txt,i},e_{sep,i},e_{img,i}$ correspondingly represent the special [CLS] token embedding, text embedding,  [SEP] token embedding and image embedding for the i-th item in user behavior sequence. N denote the number of items in user behavior sequence.

\subsection{3.4 Shared-bottom Transformer Layer}
User behavior sequence consisting of items bought by user in time series is used to represent user's behavior preference on items. Other features such as user profile features(e.g.user's age, gender and so on), items' other features excluding image and text information (e.g.item weight, item price, item location), cross features and context features are used. We concatenate all the above-mentioned other features, user behavior sequence features and target item features as the inputs of the multi-task frame. 

Since the inputs features contain sequential data, we choose multi-head transformer encoder to consume these features, which is able to learn explicit and effective sequential representation of data. The structure of transformer encoder is shown in the Figure \ref{camenn-overview}. In addition to the fact that transformer has been proved to be natural fit for sequential data \cite{Vaswani:2017:Attention}, recent application \cite{qiwei:2019:bst} shows that transformer encoder is able to model well high-order interaction between features in recommender system.

\subsection{3.5 Gating Network}
Inspired by the work of Zhen et al \cite{zhen:2020:mose}, we apply softmax gating networks to learn the weight of each transformer expert output and choose only a subset of experts conditioned on the input data for the task. The gating network can model the relationships of tasks by use different gates to separate the overlaps between tasks. For less related tasks, smaller weights will be given to sharing experts which means tasks will try to utilize different more independent experts instead. 

\subsection{3.6 Mixture of Transformer Experts (MoE) Layer}

Developed from MoE model which was introduced by Robert et al \cite{robert1991moemodel} as ensemble learning approach for multiple individual models, Eigen et al \cite{david:2014:moe} use the same structure as the MoE model turning it into MoE layer which can consume the previous layer outputs as inputs. 

We use transfomer encoders as the experts of the MoE layer to handle sequential data as well as model better features interaction. Different aspects of each task can be learned by each expert. Coordinating with gating network, MoE layer can achieve conditional computation by activating only a subset of experts for a task. 

\subsection{3.7 Task Tower Layer}
To decouple the optimization of multiple tasks, each transformer encoder tower is used for each task. The tower layer is also able to learn task-specific information.\\

\subsection{3.8 Three tasks}

Given T tasks, the above mentioned procedures for the k-th task can be formulated as:

\begin{equation}
\begin{split}
y^k =& f_{TSR}^k(X^k)\\
X^k =& \sum_{j=1}^{M} f_{TSR}( g_j^{k}(E_{Input}). f_{TSR}(X))\\
X =& f_{TSR}(E_{Input}) \\
\end{split}
\end{equation}
where $g^{k}(x) = Softmax(x)$. And $f_{TSR}$ was denoted as the transformer encoder. M is the number of Transformer Experts.

\begin{itemize}
\item \textbf{Task1: Image to Text Alignment(ITA) task}\\
As above-mentioned incorrect or vague text data issue, we exploit to use image data to align the vague text data. In this task, we use the output of special token [CLS] as the input of a binary classifier to implement the prediction whether the text data complies in the image data. For a training positive example, we choose the image feature from the item, and the text feature from the same item. On the contrary, in a negative case, the text feature is chosen randomly from other items.Binary cross-entropy loss is used to optimize the ITA objects.

\item \textbf{Task2: Text to Image Alignment(TIA) task}\\
On the other side, we implement TIA task to match the vague item image by text data. Similar to ITA task, we fed the output of special token [CLS] into a binary classifier. The positive example keep the same as the ITA task when the image and text are from the same item. For the negative training data, we randomly choose image feature from the other items when the text feature from the correct item. As a binary classification prediction, we apply binary cross-entropy loss as the objective loss function:

\item \textbf{Task3: Conversion Rate (CVR) prediction task}\\
Our main task is to predict the customers CVR which indicate whether the user would purchase the target item. For the training positive example, the target item has been bought by the user. On the contrast, the negative is the user do not buy the item. CVR prediction is also a binary classification problem. So we use the binary cross-entropy loss for the object optimization.

\end{itemize}

\section{4  Experiments}
In this section, we conduct experiments on real industry datasets and describe the main results.
\subsection{4.1 Experimental Settings}

\noindent\textbf{Dataset} : 

Our real world dataset is collected from a leading online intra-city retailing platform that mainly operates in China. Information and entities that would reveal the identities of either a shop or a consumer have been carefully removed. 

We adopt real data from our platform database, which contains data from ten millions of users and properties information from two hundreds millions of commodities. User data mainly contains user profile and user behavior information. As for the commodities' properties, we mainly parse the item image, item name and item text description. For CVR prediction task, we collect 1,257,642 positive samples and 5,030,568 negative samples. Since our dataset is from real world, typos mistakes textural data and blur image data are common. Therefore our CameNN model is designed to work in this scenario. For all three tasks, we use 75\% of total dataset for training propose and 25\% on testing dataset. \\

\noindent\textbf{Evaluation Tasks and Metrics} : 

We introduce three tasks: ITA task, TIA task and Conversion Rate (CVR) prediction task to evaluate our CameNN model. All data above-mentioned are used for training and validating these three tasks.

Since ITA task and TIA task are matching tasks, we use Accuracy to access the performance on these two matching tasks. As for the Conversion Rate (CVR) prediction task which the network is trained as a binary classification estimator, we choose Area under the ROC Curve (AUC) as the indicator to measure its predictive qualities. \\

\begin{table}[t]
\centering
\begin{tabular}{l|l}
  \hline\hline
    Methods & Offline AUC(mean$\pm$std) \\\hline
    FM    	& 0.7032$\pm$0.00081\\
    FFM   & 0.7001$\pm$0.00069  \\ 
    Wide \& Deep   & 0.7212$\pm$0.00048  \\  
    DeepFM   & 0.7284$\pm$0.00087  \\  
    AFM    & 0.7230$\pm$0.00056 \\
    xDeepFM & 0.7289$\pm$0.00064 \\
    DIN & 0.7336$\pm$0.00045 \\
    BST & 0.7412$\pm$0.00038 \\\hline
    \bf CameNN & \bf 0.7549$\pm$0.00034\\\hline
\end{tabular}
\caption{Experimental CVR Prediction results of CameNN and baseline models}
\label{Table1}
\end{table}

\noindent\textbf{Implementation Details} : 

The dataset is split into two parts, with 75\% being used for training and 25\% for the testing

We use the chinese version pre-trained BERT with 12-layer ,768-hidden, 12-heads and 110M parameters to generate the text token embeddings. For the image patch feature generating, pre-trained from imagenet dataset InceptionV4 model with num\_classes equal to 1001 was adopted. We set 50 as the maximum text sequence length and 9 (3*3 patches) as the maximum patch sequence length. Our experiments are conducted with Tensorflow and trained with 4 NVIDIA TITAN-V GPUs. 
For the transformer encoders used in CameNN, the encoders contain 8 heads and 1 transformer encoder block. All experimental models are trained with the Adam \cite{Kingma:2014:Adam} optimizer with a learning rate of $4 \cdot 10^{-4}$ with $\beta1$= 0.95, $\beta2$ = 0.999 and weight decay of 1e-4. Early stopping method is also implemented to avoid over-fitting. 

\subsection{4.2 Evaluations with baselines}
To explore the performance of our model on CVR prediction, we implemented state-of-the-art deep learning models in this area as baselines on the same dataset.

\begin{itemize}

\item FM \cite{Steffen:2010:Factorization-Machines} : FM takes advantage of factorization mechanism to model second-order feature interactions. 

\item FFM  \cite{Yuchin:2016:Field-aware-Factorization-Machines} : FFM models fine-grained interactions between features from different fields 

\item Wide\&Deep \cite{ Cheng2016:Wide-deep} : Wide\&Deep uses jointly training feed-forward neural networks with embeddings and linear model with feature transformations. 

\item  DeepFM  \cite{Guo:2017:DeepFM}: DeepFM combines the power of factorization machines for recommendation and deep learning for feature learning in the same neural network.

\item AFM \cite{Xiao:2017:Attentional-factorization-machines}: AFM extends FM by using attention mechanism to determine the different importance of second-order combinatiorial features to capture second-order feature interaction. 

\item xDeepFM \cite{Lian:2018:xDeepFM}: xDeepFM uses compressed Interaction Network to take outer product of stacked feature matrix at vector-wise level. 

\item DIN \cite{guorui:2018:din}: With attention mechanism to deal with the users' behavior sequences, DIN model try to capture different similarities between previously clicked items and target item.

\item BST \cite{qiwei:2019:bst}: BST takes advantage of Transformer's powerful abilities to capture sequential relations so that model be able to learn deeper representation for items in user' behavior sequences. 

\end{itemize}  

Table.\ref{Table1} indicates that comparison of CameNN against all baselines above-mentioned shows that our model achieve the best performance with respect to the chosen metric AUC in CVR prediction task.

\subsection{4.3 Ablation Studies} 

\textbf{1) ITA and TIA tasks study}

FashionBERT \cite{dehong:2020:fashionbert} introduces an innovative method to implement text and image matching and cross-modal retrieval, consisting of pre-train BERT \cite{jacob:2018:bert} as the matching backbone network and an adaptive loss to trade off multi-tasks.

\begin{table}[t]
\centering
\begin{tabular}{l|l|l}
  \hline\hline
    Methods & TIA Accuracy & ITA Accuracy\\\hline
    FashionBERT & 85.13\% & 85.71\% \\
    \bf CameNN & \bf 85.08\% &  \bf 86.23\% \\\hline
\end{tabular}
\caption{Experimental TIA and ITA Task results of CameNN and FashionBERT}
\label{Table2}
\end{table}

\begin{figure}[t]
\centering
\includegraphics[width=0.7\columnwidth]{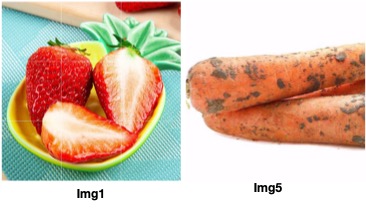} 
\caption{\\\textbf{text1}: Juicy red sweet berries 500g. Locally farm grown. No farm chemical used\\\textbf{text5}: Fresh carrots from local farm. No chemical.}.
\label{pic}
\end{figure}

To evaluate the performance of CameNN on ITA and TIA task, we set series of experiments comparing with FashionBERT. As descibe above, we use accuracy as the metric to access the performance on ITA and TIA tasks. Table.2 shows that CameNN is able to achieve evenly matched performance as FashionBERT on image and text alignment tasks.In addition, CameNN is more flexible for less related multi-tasks.\\

\noindent \textbf{2) Visualization of embedding similarity after alignment tasks}

\begin{figure}[t]
\centering
\includegraphics[width=0.7\columnwidth]{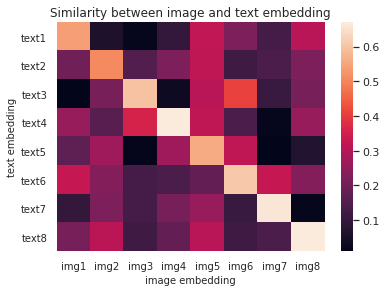} 
\caption{Text and image embedding similarity before TIA and ITA Task}.
\label{heatmap_before}
\end{figure}

\begin{figure}[t]
\centering
\includegraphics[width=0.7\columnwidth]{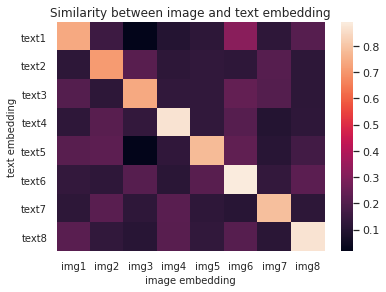} 
\caption{Text and image embedding similarity after TIA and ITA Task}.
\label{heatmap_after}
\end{figure}

\begin{table}[t]
\centering
\begin{tabular}{l|l}
  \hline\hline
    Methods & CVR Gain\\\hline
    xDeepFM &   -4.20\%$\pm$0.012\% \\
    DIN &   -2.76\%$\pm$0.027\% \\
    BST &   -0.05\%$\pm$0.080\% \\
    \bf CameNN & \bf 0.63\%$\pm$0.014\%\\\hline
\end{tabular}
\caption{Practical CVR gain results of CameNN and baseline models}
\label{pracitical-gain}
\end{table}

\begin{table}[t]
\centering
\begin{tabular}{l|l}
  \hline\hline
    Methods & Offline AUC(mean$\pm$std)\\\hline
    MMoE frame & 0.7203 $\pm$0.00056 \\
    MoSE frame & 0.7516 $\pm$0.00041  \\
    \bf CameNN & \bf 0.7549$\pm$0.00034 \\\hline
\end{tabular}
\caption{Experimental CVR Task Prediction results of CameNN and FashionBERT}
\label{multitask-study}
\end{table}

Figure.\ref{pic} shows the item 1 which is a typical vague text information data, the image of strawberries is explicit and clear while the corresponding text is ambiguous with only berries to define the commodity. On the other hand, the item 5 in Figure. \ref{pic}, the text data describes unequivocally about the properties of carrots while the image is not determinable.

To visualize the alignment tasks impacts on item text embedding and image embedding, we choose 8 items as samples. Each item consists of two parts: text embedding and image embedding. We calculate the similarity for each item between item image embedding and corresponding text embedding before and after the ITA and TIA tasks.In the heatmaps illustrated by Figure.\ref{heatmap_before} and Figure.\ref{heatmap_after}, the similarity between the text and image embedding from the same item increase after the implementation of alignment tasks.\\

\noindent \textbf{3) Multi-task frame study}

\noindent \textbf{MMoE} \cite{Ma:2018:mmoe}: MMoE adapts Mixture-of-Experts(MoE) and softmax gating network to implement multi-task learning across tasks by sharing subsets of experts. All the experts are fully connect layer with ReLU activation.\\
\noindent \textbf{MoSE} \cite{zhen:2020:mose}: MoSE applies the same frame as MMoE and replace the ReLU experts and ReLU shared-bottom layer with LSTM layer \cite{Hochreiter:1997:LSTM} to capture sequential features. 

In CameNN we use transformer encoder as the functional layer of the experts and the shared-bottom layer. Comparing with MMoE and MoSE, the Table. \ref{multitask-study} shows that CameNN outperform both MMoE and MoSE on CVR prediction task.

\subsection{4.5 Industry Applications} 

To demonstrate the effective of CameNN in practical application, we use the first seven days data as the training data, the eighth day as the testing data. Then we use online A/B test to evaluate only three baseline models that achieve top three offline AUC for controlling the impacts on real-world business and CameNN performance on online CVR Gain. As illustrated in Table. \ref{pracitical-gain}, CameNN help to increase the online CVR by 0.63\% while other baselines do not improve the performance. 

\section{5  Conclusion}
In this paper, we propose a multi-task recommendation model named Cross-modal Alignment with mixture experts Neural Network (CameNN) to model CVR prediction and image-text alignment. CameNN achieves significant improvement on CVR prediction in intral-city retail industry. Specially, the ITA and TIA tasks contribute to rectify incorrect image and text data of items, which reduce the misleading noise of dirty data. We adapt transformer encoder as the basic block of shared-bottom layer, mixture of experts layer and task tower layers to handle sequential data. Benefiting from the softmax gate networks which achieve conditional computation by activating only a subset of experts for a task, We show a successful and efficient large-scale online application of CameNN to improve CVR of items recommendation. In the future, we will try to explore more potential tasks in this model. 

\bibliography{camenn.bib}
\end{document}